# Semantic Photometric Bundle Adjustment on Natural Sequences

Rui Zhu, Chaoyang Wang, Chen-Hsuan Lin, Ziyan Wang, Simon Lucey
The Robotics Institute, Carnegie Mellon University
{rz1, chaoyanw, chenhsul, ziyanw1}@andrew.cmu.edu, slucey@cs.cmu.edu

## Abstract

*The problem of obtaining dense reconstruction of an object in a natural sequence of images has been long studied in computer vision. Classically this problem has been solved through the application of bundle adjustment (BA). More recently, excellent results have been attained through the application of photometric bundle adjustment (PBA) methods – which directly minimize the photometric error across frames. A fundamental drawback to BA & PBA, however, is: (i) their reliance on having to view all points on the object, and (ii) for the object surface to be well textured. To circumvent these limitations we propose semantic PBA which incorporates a 3D object prior, obtained through deep learning, within the photometric bundle adjustment problem. We demonstrate state of the art performance in comparison to leading methods for object reconstruction across numerous natural sequences.*

## 1. Introduction

In this paper we are primarily concerned with the goal of obtaining dense 3D object reconstructions from short natural image sequences. One obvious strategy is to employ classical bundle adjustment (BA) [18] across the sequence where we can simultaneously recover camera pose and 3D points. Although reliable, this strategy is problematic as it: (i) can only recover 3D points if they are observed in the image sequence, and (ii) the density of the reconstruction is dependent on how textured the surface of the object is across the image sequence. Recently, photometric extensions to bundle adjustment have been proposed [1, 4] that directly minimize the photometric consistency between frames with respect to pose and 3D points. Borrowing upon the terminology of [1] we shall refer to these methods collectively herein as photometric bundle adjustment (PBA). PBA has recently proved advantageous over classical BA for problems where dense reconstructions are required due to their ability directly minimize for photometric consistency. Even with these recent innovations PBA is still, however, fundamentally limited in performance by (i) and (ii).

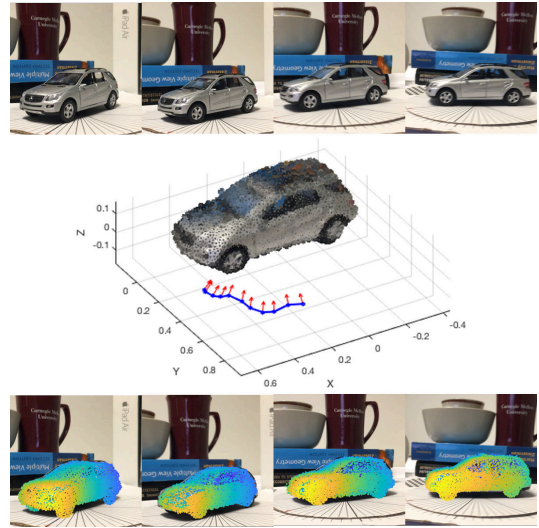

Figure 1: **Overview of the proposed method for semantic object-centric photometric bundle adjustment.** Given a sequence of images with small motion (top row), we recover a full 3D dense shape of the object, as well as the camera poses w.r.t. the global coordinate system (middle row). Our method enables reprojection to the image plane with the estimated cameras (bottom row) to optimize for the photometric consistency as well as silhouette and depth constraints.

There have been numerous efforts within the computer vision community to bring semantic prior into the task of object/scene 3D reconstruction. Convolutional Neural Networks (CNN) are proving remarkably useful for this task when one is provided with scene [17, 24] or object category [25, 7] specific labels & priors. A powerful characteristic of these semantic CNN methods is their ability to circumvent the fundamental limitations of (i) and (ii). For example, [25, 7, 21, 6] offer strategies for inferring a dense 3D reconstruction of an object from a single image even when a substantial portion of the projected 3D points are self-occluded. More recently, semantic CNN strategies have been proposed that attempt to incorporate multiple frames [3, 9]. Most of these previous efforts have been trying to attack the problem of 3D reconstruction as a supervised learning problem – where geometry is largely treated as a label to be predicted.

Although attractive in its simplicity this strategy has some inherent drawbacks. First, these geometric labels can be problematic to obtain – hand labeling can be error prone, and rendering can lack the necessary realism. Second, the predicted labels from these network models do not adhere to geometric constraints – such as photometric consistency – making the results unreliable. Recently, the application of geometric constraints within the offline CNN learning process has been entertained including reprojected silhouette matching [22, 25], depth matching [14], and even photometric consistency [19, 13, 12]. A fundamental problem with these emerging methods, however, is that the geometric constraints are not enforced at test time – dramatically reducing their effectiveness.

Given these concerns, we argue that instead of incorporating geometric constraints into semantic CNN strategies offline, one should instead incorporate object semantics within the PBA pipeline. As we demonstrate in Fig. 1 and our results section this strategy gives the best of both worlds – semantic knowledge of the object with photometric consistency. In this paper, we propose an enhanced semantic PBA method which works on natural sequences as the classic PBA does, and give extensive evaluations on both synthetic and natural sequence domains. We summarize our contributions as follows:

- We provide the first approach of its kind (to our knowledge) for semantic object-centric PBA on natural sequences – which gives the global 6DoF camera poses of each frame and the dense 3D shape, with PBA-like accuracy but denser depth maps.

- We systematically evaluate the local optimality of our proposed optimization pipeline, as well as our enhanced objective which takes use of classic PBA results as off-the-shelf initialization or regularizer in our method.

- We collect a new dataset for the task of object-centric shape reconstruction, consisting of rendered sequences of full ground truth in cameras, depth maps, and the shape in canonical pose, as well as natural sequences annotated with ShapeNet [2] models, making the dataset feasible for evaluating both PBA methods with camera estimations [11, 8], and methods which only recover aligned shapes or depth maps [20, 13, 9] by end-to-end learning on ShapeNet .

## 1.1. Related Work

**Photometric Bundle Adjustment:** Photometric bundle adjustment (PBA) which is an optimization based method sitting entirely upon the visual cue of photometric consistency across all input frames [5, 15, 1, 11]. In PBA the shape is recovered by jointly optimizing for depth maps corresponding to the visible pixels in template frames [5, 15, 1], as well as camera motion. As a result the formulation of classic PBA is solely to recover the geometry of the scene, completely agnostic to semantics of the scene/object. Some works in PBA aims for small motion videos in particular [11, 10].

**Shape Reconstructing with Deep Learning:** As previously mentioned, early deep learning methods [7, 21, 6, 3, 9] solve the task of object-centric (object shape only) reconstruction with supervision from shape labels. Recently, an emerging school of thought seeks to bring in geometry back to the task, including reprojected silhouette matching [22, 25], depth matching [14], and even photometric consistency [19, 13, 12]. One issue of most of these methods is they assume known cameras, in global frame. This is in fact a too strong assumption to hold for natural sequences where global camera poses are not readily availabe. While some others do not account for camera motion at all [20, 3], which creates a gap from classic PBA where camera motions are instead the direct output.

**Semantic PBA:** Recent work of Zhu *et al*. [26] also proposed to apply shape priors within PBA for 3D object reconstruction. In spite of the similarity in the formulation of the problem, Zhu *et al*.'s approach was problematic in a number of ways. First, the performance of Zhu *et al*. relies heavily on the initialization point given by CNN pose/style predictors trained predominantly on rendered images, which is suspicious for its reliability on natural sequences. Instead, we utilize a more reliable source – relative camera pose from PBA for initialization. Second, due to the limitations of the method, such as weak perspective camera model assumption and unreliable initialization source, Zhu *et al*.'s evaluation was restricted to rendered sequence. Thus they did not conduct quantitative comparisons between actual PBA methods [11, 8] w.r.t. camera pose error or depth error on real world sequences; while we give an extensive evaluation of our method on real sequences. Third, Zhu *et al*. did not give a proper analysis of the characteristics of their objective function, which results in using inadequate optimization techniques for their approach; In this paper, through inspecting the property of different cost functions, we propose a more robust and efficient optimization pipeline.

**Notation:** Vectors are represented with lower-case bold font (*e.g.* **a**). Matrices are in upper-case bold (*e.g.* **M**) while scalars are lower-cased (*e.g. a*). Italicized upper-cased characters represent sets (*e.g. S*). To denote the $l^{th}$ sample in a set (*e.g.* images, shapes), we use subscript (*e.g.* **M**$_l$). Calligraphic symbols (*e.g.* $\pi$) denote functions. Images are defined as sampling function over the pixel coordinates, *i.e.* $\mathcal{I}(\mathbf{u}) : \mathbb{R}^2 \to \mathbb{R}^3$.

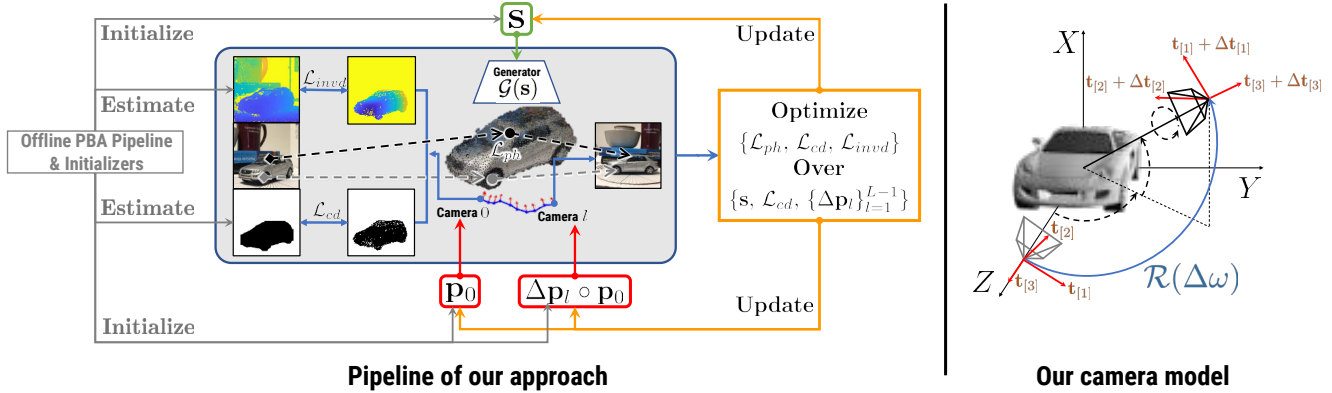

Figure 2: **(Left) Pipeline of the optimization.** Given an input sequence, we first run it through offline PBA pipeline to provide rough estimation for the depth maps as future depth map constraint, and camera motion as initialization to our camera motion parameters. Our style & pose initializer is also called to initialize the global pose of the template frame, as well as the style vector. Starting from the initialization, we optimize all variables until convergence through PBA-like pipeline with combined objective from photometric consistency $\mathcal{L}_{ph}$, silhouette matching error $\mathcal{L}_{cd}$ and inverse depth error $\mathcal{L}_{invd}$. **(Right) The perspective camera model**. We adopt the camera model in Blender$^{\text{TM}}$ by centering the object in the world frame, and positioning the camera with identity pose (in grey) along the Z axis, looking at the origin point. We show in this figure camera transformation from the identity camera can be parameterized with exponential twist $\Delta\omega$ (solid blue arrow) and translation $\Delta\mathbf{t}$ (red arrows) w.r.t. the camera local axes.

## 2. Approach

### 2.1. Preliminary

**Camera model** We assume a perspective camera with known intrinsics $\mathbf{K}$. The camera extrinsics are parameterized as concatenation of exponential coordinates (also known as twist) of rotation and translation vector: $\mathbf{p} = [\omega; \mathbf{t}] \in \mathbb{R}^6$. The camera projection function is written as

$$\pi(\mathbf{x}; \mathbf{p}) = \mathbf{K}(\mathcal{R}(\omega)\mathbf{x} + \mathbf{t}). \quad (1)$$

Given a short window of $L$ frames as in PBA [11, 1], we define the first frame as the target frame (frame 0), and the subsequent $L-1$ frames as source frames. The relative camera pose between the target frame and source frame is denoted as $\Delta\mathbf{p}_l = [\Delta\omega_l; \Delta\mathbf{t}_l]$. The global pose is thus computed as a composing of relative pose of the source frame and the global pose of the target frame:

$$\mathbf{p}_l = [\Delta\omega_l \circ \omega_0; \Delta\mathbf{t}_l + \mathbf{t}_0]. \quad (2)$$

We define the pose composition rule as:

$$\mathcal{R}(\Delta\omega \circ \omega) = \mathcal{R}(\Delta\omega)\mathcal{R}(\omega), \quad \Delta\mathbf{t} \circ \mathbf{t} = \Delta\mathbf{t} + \mathbf{t}. \quad (3)$$

The reprojection of one point $\mathbf{x}$ onto frame $l$ with the corresponding global pose is framed as sampling the image with reprojected pixel location $\mathcal{I}_l(\pi(\mathbf{x}_j; \mathbf{p}_l))$.

**Reprojection with pseudo-raytracing** Reprojection from a point set $X$ given camera pose $\mathbf{p}$ can be viewed as first reasoning the visible part of the point set with a masking function $X_{\mathbf{p}} := \mathcal{M}(X; \mathbf{p})$ where $X_{\mathbf{p}} = \{\mathbf{x}_j\}_{j=1}^{M_{\mathbf{p}}}$. The mask function is implemented as pseudo-raytracing [14] by projecting the points to a enlarged inverse depth plane by a factor $U$, and then perform max pooling in a neighbourhood of $U \times U$ to figure out the visible point with biggest inverse depth. By doing so the mask function gives both the indices of the $M_{\mathbf{p}}$ visible points, and an inverse depth map.

### 2.2. Overview

Our method takes in a RGB sequence taken by a calibrated camera moving around an object. The category(*e.g.* cars, airplanes, chairs) of the object is assumed to be known, and we have a rich repository of aligned CAD models (*e.g.* ShapeNet [2]) for this category. We define the world coordinate system as one attached to the objects as chosen by the CAD dataset, and the calibrated perspective camera model is parametrized by full 6DoF rotation and translation (see Fig.2). The goal of our method is to recover the full 3D shape of the object in the world frame, as well as the 6DoF parameters of the camera pose of each frame.

We adopt the category-specific shape prior from Zhu et.al [26] to learn a shape space from the repository of ShapeNet [2] CAD models. We use dense point cloud as the shape parameterization in our work, considering learning shape space of point clouds has been made possible by several works [6, 16, 26]. The shape prior is a learned category-specific point cloud generator, written as a function of a **style vector** $\mathbf{s} \in \mathbb{R}^S$ which represents the sub-category object style. The output of the shape prior is the set of gener-ated points defined as $X := \{\mathbf{x}_i\}_{i=1}^N = \mathcal{G}(\mathbf{s})$. For the 6DoF poses of the total $L$ frames, we break the pose parameters into two sets: the global camera pose of the target frame $\mathbf{p_0} \in \mathbb{R}^6$, and the relative camera pose between each source frame and the corresponding target frame $\{\boldsymbol{\Delta}\mathbf{p}_l \in \mathbb{R}^6\}_{l=1}^{L-1}$.

The overall pipeline of our method is illustrated in Fig. 2. We formulate the inference of the style vector and camera pose parameters as optimization steps with geometric objectives. The parameters are initialized with an off-the-shelf initialization pipeline (see 2.4), and the optimization steps are taken to minimize this objective over the parameter space. In this paper, we propose to take advantage of the cheap and rough outputs from other methods to regularize our optimization procedure. For each frame, we get cheap segmentation masks (silhouettes) from recent state-of-the-art instance segmentation method FCIS [23]. Considering traditional PBA methods gives as results semi-dense inverse depth and camera motion, we also borrow the readily available although error-prone outputs from PBA pipelines (*e.g.* openMVS [8]) to add another inverse depth loss with the estimated depth. We also take advantage of their camera motion estimation to initialize the relative camera pose of each source frame w.r.t. its target frame.

### 2.3. Optimization Objective

**Photometric Consistency** The basic objective is formulated as the photometric consistency between the corresponding pixel pairs from the target frame and each source frame. Classic PBA methods usually track a set of sparse points through all the frames in a window, while with our formulation we are able to get dense correspondence automatically derived from reprojection. Considering that visible points may differ in each frame due to camera motion and occlusion, in this work we formulate the bi-directional photometric consistency as:

$$\mathcal{L}_{\text{ph}}(\mathbf{p}_0, \{\boldsymbol{\Delta}\mathbf{p}_l\}_{l=1}^{L-1}, \mathbf{s}) = \quad (4)$$
$$\sum_{l=1}^{L-1} \left[ \sum_{j=1}^{M_{\mathbf{p}_0}} \mathcal{L}_{\delta_1}\Big(\mathcal{I}_0\big(\pi(\mathbf{x}_j; \mathbf{p}_0)\big) - \mathcal{I}_l\big(\pi(\mathbf{x}_j; \mathbf{p}_0 \circ \boldsymbol{\Delta}\mathbf{p}_l)\big)\Big) \right.$$
$$\left. + \sum_{k=1}^{M_{\mathbf{p}_l}} \mathcal{L}_{\delta_1}\Big(\mathcal{I}_l\big(\pi(\mathbf{x}_k; \mathbf{p}_0 \circ \boldsymbol{\Delta}\mathbf{p}_l)\big) - \mathcal{I}_0\big(\pi(\mathbf{x}_k; \mathbf{p}_0)\big)\Big) \right]$$

where $\mathcal{L}_\delta(\cdot)$ is the Huber loss.

**Silhouette Error and Inverse Depth Error as Extra Constraints** In Zhu *et al.* [26] the silhouette error is utilized as an extra constraint in the objective. We adopt the same objective but with the estimated silhouette with FCIS [23] that produce instance segmentation. We will show later that this constraint is still effective although the masks are error-prone. We write the silhouette distance for frame $l$ as the 2D Chamfer distance between the set of pixel locations $U_{l1}$ in-side the rough silhouette, and the ones projected down from our camera model $U_{l2}$.

$$\mathcal{L}_{\text{cd}}(\mathbf{p}_0, \{\boldsymbol{\Delta}\mathbf{p}_l\}_{l=1}^{L-1}, \mathbf{s}) = \quad (5)$$
$$\frac{1}{L}\sum_{l=0}^{L-1}\Big(\sum_{\mathbf{u}_k \in U_{l1}} \min_{\mathbf{u}_j \in U_{l2}} ||\mathbf{u}_k - \mathbf{u}_j||_2^2$$
$$+ \sum_{\mathbf{u}_j \in U_{l2}} \min_{\mathbf{u}_k \in U_{l1}} ||\mathbf{u}_j - \mathbf{u}_k||_2^2\Big).$$

Finally, apart from cheap camera motion, we are able to get semi-dense depth map for each frame from the off-the-shelf PBA pipeline. In this case we further formulate the extra objective term of inverse depth error as:

$$\mathcal{L}_{\text{invd}}(\mathbf{p}_0, \{\boldsymbol{\Delta}\mathbf{p}_l\}_{l=1}^{L-1}, \mathbf{s}) = \frac{1}{L}\sum_{l=0}^{L-1}\mathcal{L}_{\delta_2}(\mathbf{d}'_l - \alpha\mathbf{d}_l) \quad (6)$$

where $\mathbf{d}_l$ is given by our reprojection module. Note that $\alpha$ here should be updated on the fly. Considering we will be getting confident camera poses for all frames in a few optimization steps, $\alpha$ can be robustly solved by finding a scale that best aligns the estimated camera poses with the ones from the offline estimator. Specifically, we can solve for an $\alpha_l$ for each source frame $l$:

$$\begin{cases} \mathcal{R}_0(\omega_0)\mathbf{x} + \mathbf{t}_0 = \alpha_l\Big(\mathbf{R}'_0\mathbf{x} + \mathbf{t}'_0\Big) \\ \mathcal{R}_l(\omega_l)\mathbf{x} + \mathbf{t}_l = \alpha_l\Big(\mathbf{R}'_l\mathbf{x} + \mathbf{t}'_l\Big). \end{cases} \quad (7)$$

The solution for $\alpha$ is average over all $\{\alpha_l\}_{l=1}^{L-1}$:

$$\alpha = \frac{1}{L-1}\sum_{l=1}^{L-1}\arg\min_\alpha ||\mathcal{R}_l(\omega_l)\mathbf{x} + \mathbf{t}_l \quad (8)$$
$$- \mathbf{R}'_l\mathbf{R}_0^T(\mathcal{R}_0(\omega_0)\mathbf{x} + \mathbf{t}_0) - \alpha(\mathbf{t}'_l - \mathbf{R}'_l\mathbf{R}'^T_0\mathbf{t}'_0)||_2^2.$$

The combined objective is given by:

$$\mathcal{L} = \mathcal{L}_{\text{ph}} + \lambda_1\mathcal{L}_{\text{cd}} + \lambda_2\mathcal{L}_{\text{invd}}, \quad (9)$$

where ablative study about the weight factors $\lambda_1$ and $\lambda_2$ is included the Appendix.

### 2.4. Initialization

**Style and Pose Initialization** We improve upon existing learning-based pipeline to provide initialization for style and template frame camera pose. For style, unlike [25, 26] where a single-image based regressor is adopted, we use a recurrent network architecture to leverage the sequential information in an effort to alleviate ambiguity in style from a single viewpoint. Details about the architecture of our regressor and the training process as well as dataset are included in the Appendix. To generate more accurate style

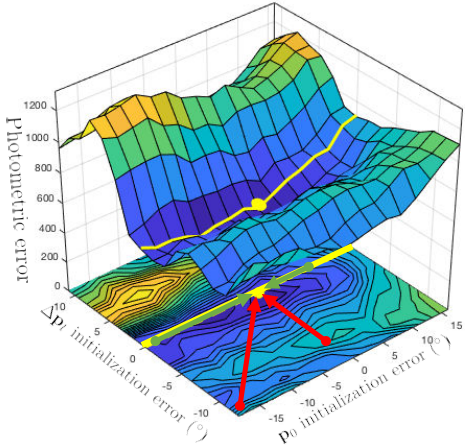

Figure 3: **Local cost surface of $\mathbf{p}_0$ and $\Delta \mathbf{p}_l$.** The yellow dot marks the optimum while the yellow line shows the search space for $\mathbf{p}_0$ if $\Delta \mathbf{p}_l$ is initialized with offline camera motion.

vectors, we exploit cheap silhouette from Li *et al.* [23] to mask off the background.

To find a coarse pose initialization, we first utilize Blender$^{TM}$ to render templates under varying camera poses. After that we retrieve a coarse pose by finding one template which has the maximum IoU with the target silhouette.

**Camera Motion Initialization** One observation of ours is, the photometric consistency error is problematic in our optimization of the objective. On the one hand $\mathcal{L}_{\text{ph}}$ is locally bi-linear with Gauss-Newton solvers, where in traditional PBA the template term is fixed so that the residual is linear and the problem is locally convex. The problem is worsened in the way the variables are initialized as in Zhu *et al.* [26], where $\{\Delta \mathbf{p}_l\}_{l=1}^{L-1}$ are initialized to all zeros when we have a poorly-initialized $\mathbf{p}_0$. To illustrate this we show in Fig. 3 by plotting the cost surface over local perturbation of $\mathbf{p}_0$ and $\Delta \mathbf{p}_l$. We show that the cost surface is highly non-convex between the initialization and the optimum if both variables are initialized far from the ground truth (red arrows). However we show with the camera motion parameters initialized correctly, the search space of $\mathbf{p}_0$ is constrained to the yellow line with better curvature for better convergence (green arrows).

Inspired by this observation, at the beginning of our pipeline we run an off-the-shelf PBA pipeline (*e.g.* [8]) to acquire camera poses $\{\mathbf{p}'_l\}_{l=0}^{L-1}$ for every frame, as well as a semi-dense point cloud $X'$. By formulation we have the following relation between our camera model (left) and the off-the-shelf estimator (right):

$$\mathcal{R}_l(\omega_l)\mathbf{x} + \mathbf{t}_l = \alpha\Big(\mathcal{R}'_l(\omega'_l)\mathbf{x}' + \mathbf{t}'_l\Big). \tag{10}$$

Unfortunately no correspondence between $\mathbf{x}$ and $\mathbf{x}'$ is available to give an accurate estimation of the scale factor $\alpha$. Instead we seek to bring the estimation of $\alpha$ in the loop by aligning the estimated inverse map $\mathbf{d}'_0$ to our reprojected inverse depth $\mathbf{d}_0$:

$$\arg\min_{\alpha} ||\mathbf{d}'_0 - \alpha \mathbf{d}_0||_2^2. \tag{11}$$

The camera motion is then initialized by solving:

$$\begin{cases} \mathcal{R}_0(\omega_0)\mathbf{x} + \mathbf{t}_0 = \alpha\Big(\mathbf{R}'_0\mathbf{x} + \mathbf{t}'_0\Big) \\ \Delta\mathcal{R}_l(\Delta\omega_l)\mathcal{R}_0(\omega_0)\mathbf{x} + \Delta\mathbf{t}_l + \mathbf{t}_0 = \alpha\Big(\mathbf{R}'_l\mathbf{x} + \mathbf{t}'_l\Big) \end{cases} \tag{12}$$

where the solution for $\Delta\omega_l$ and $\Delta\mathbf{t}_l$ is:

$$\begin{cases} \Delta\mathcal{R}_l(\Delta\omega_l) = \mathbf{R}'_l\mathbf{R}'^T_0 \\ \Delta\mathbf{t}_l = \mathbf{R}'_l\mathbf{R}'^T_0\mathbf{t}_0 + \alpha\Big(\mathbf{t}'_1 - \mathbf{R}'_l\mathbf{R}'^T_0\mathbf{t}'_0\Big). \end{cases} \tag{13}$$

We use Equ. 13 for initializing the camera motion parameters before the optimization steps.

### 2.5. Optimization Pipeline

Given reasonable initialization from the above steps, we solve our objective with gradient-descent based methods. Particularly we found off-the-shelf L-BFGS solver gives efficient solution to our problem. We summarize our optimization pipeline in Algorithm 1.

---

**Algorithm 1** Optimization of the objective

1: **procedure** $\mathcal{L}(\mathbf{p}_0, \{\Delta\mathbf{p}_l\}_{l=1}^{L-1}, \mathbf{s})$
2:     $\mathbf{p}_0, \mathbf{s} \leftarrow \text{Initializer}(\mathcal{I}_0)$
3:     $\alpha, \{\Delta\mathbf{p}_l\}_{l=1}^{L-1}, \{\mathbf{d}'_l\}_{l=0}^{L-1} \leftarrow \text{openMVS [8], Equ. 13}$
4:     **while** $\mathcal{L} > \delta_L$ or step < maximum iterations **do**
5:         $\mathbf{p}_0 \leftarrow$ L-BFGS update on $\mathbf{p}_0$
6:         $\{\Delta\mathbf{p}_l\}_{l=1}^{L-1} \leftarrow$ L-BFGS update on $\{\Delta\mathbf{p}_l\}_{l=1}^{L-1}$
7:         $\mathbf{s} \leftarrow$ L-BFGS update on $\mathbf{s}$
8:         $\alpha \leftarrow$ Update on $\alpha$ by Equ. 8
9:     **return** $\{\mathbf{p}_0, \{\Delta\mathbf{p}_l\}_{l=1}^{L-1}, \mathbf{s}\}$

---

## 3. Evaluation

### 3.1. Data Preparation

**Rendered Data** To enable evaluation of our methods against Zhu *et al.* [26] which is only feasible on rendered domain, we follow Zhu *et al.* by rendering small baseline sequences from ShapeNet [2] cars. Please refer to Appendix for the statistics which is inherently identical to Zhu *et al.* apart from our perspective camera versus weak-perspective by Zhu *et al.*

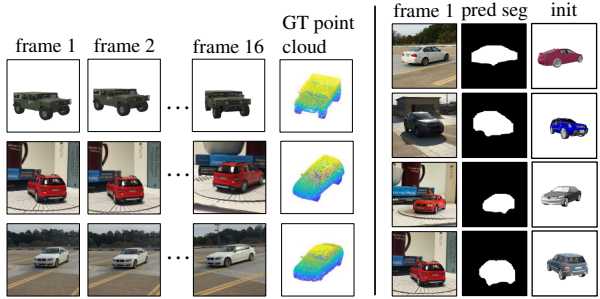

Figure 4: **(Left) Examples of our test sequences.** We show on the left three examples of rendered sequence, natural sequence of model car and real car respectively. **(Right) Visualization of style & pose initialization.** We also show some quantitative results on style and pose initialization for some natural sequences. The three columns on the right correspond to the first frame of a sequence, foreground segmentation of that first frame, and style & pose initialization viewed in rendered image, respectively.

**Natural Data** In view of the absence of object-centric and category-specific natural video dataset, we collect our own test set with a mixture of sequences from toy model cars and real cars. We carefully choose the models with ground truth CAD available in ShapeNet [2]. Each video is shot with iPhone 6s (30fps) with around 30 °of rotational motion and moderate translation, and the images are scaled down to $512 \times 512$. There are 15 videos collected from 4 toy car models and 12 videos from 6 real cars for evaluation. For each sequence, we annotate the template frame and last frame with ground truth CAD models and 6DoF pose. Samples of our test data can be found in Fig. 4.

We also visualize some qualitative results of our initialization on natural sequences in Fig. 4. Although the style retrieved from regressor is not very precise in color or details, our style regressor is able to yield style prediction that are close in 3D shape.

### 3.2. Evaluation on Rendered Sequences

In this section, we give both qualitative and quantitative results of our methods against 1) classic PBA methods of open-source openMVS [8], and Ham *et al.* [11] which is a PBA pipeline specifically optimized for small motion videos, 2) learning based methods [6, 3] which gives shapes in canonical poses.

In our experiments, we set the weights in the optimization objective as $\lambda_1 = 0.1$ with $\delta_1 = 100$ (we use unnormalized RGB values in range [0, 255]), and $\lambda_1 = 1000$ with $\delta_2 = 10$. Ablative study on different settings of the hyper-parameters is included in the Appendix.

Since we are evaluating on synthetic sequences from Blender$^{TM}$, we have access to full ground truth for both the camera pose of every frame (in world coordinate system), as well as the ground truth dense depth map and dense shape. However given our formulation is to output full 3D shape in an object-centric manner, while openMVS and Ham *et al.* give semi-dense point cloud of the entire scene by best effort, we are not able to align three outputs to the ground truth shape. Instead we choose to follow Ham *et al.* [11] to measure the depth error of the recovered points by reprojecting the shape onto the image plane with the estimated cameras. To compare with openMVS [8] and Ham *et al.* [11] which give only relative camera motion of every source frames w.r.t. to the first frame, we offer ground truth camera pose of the first frame to these two methods and find a rigid-body transformation to align their camera pose of the first frame to the ground truth. The scale ambiguity is solved by Equ. 8. By doing this, the shape and camera poses of all methods are ideally aligned to the world coordinate system, and we are able to measure the camera error by calculating the camera position error as the distance of the estimated camera center to its ground truth, and the camera orientation error as the acute angle between the estimated camera orientation and its ground truth.

We report the average error of depth maps and camera poses in Table 1 and the statistics in Fig. 6. The results show that our method achieves comparable camera error as openMVS [8], but slightly worse than Ham *et al.* [11] which is specifically optimized for small motion videos for better camera tracking. For the depth map error we achieve SLAM-like performance by outperforming both openMVS [8] and Ham *et al.* [11] in addition to much denser results thanks to the shape prior which produces full 3D shapes.

Moreover, again thanks to the shape prior, we show in Fig. 7 that we only need a few observations of the object to give confident results, even at two frames, while classic methods [8, 11] start at considerable amount of motion to perform camera tracking.

Finally we experiment on all the rendered sequences by perturbing upon their ground truth $\mathbf{p}_0$ and calculate the average $\mathbf{p}_0$ at convergence. We show in Fig. 8 that our method achieves better convergence in face of large initialization error in pose.

### 3.3. Evaluation on Natural Sequences

We evaluate our methods against others on the object-centric dataset we collected. The dataset includes sequences of a mixture of toy cars and real cars. The sequences are annotated with aligned CAD models retrieved from the ShapeNet dataset [2]. Considering it is not possible to get the 6DoF camera poses for all frames through human annotation, we evaluate on the depth error of the annotated first frame of each sequence, as well as the density of the reprojected points against the ground truth. The quantitative

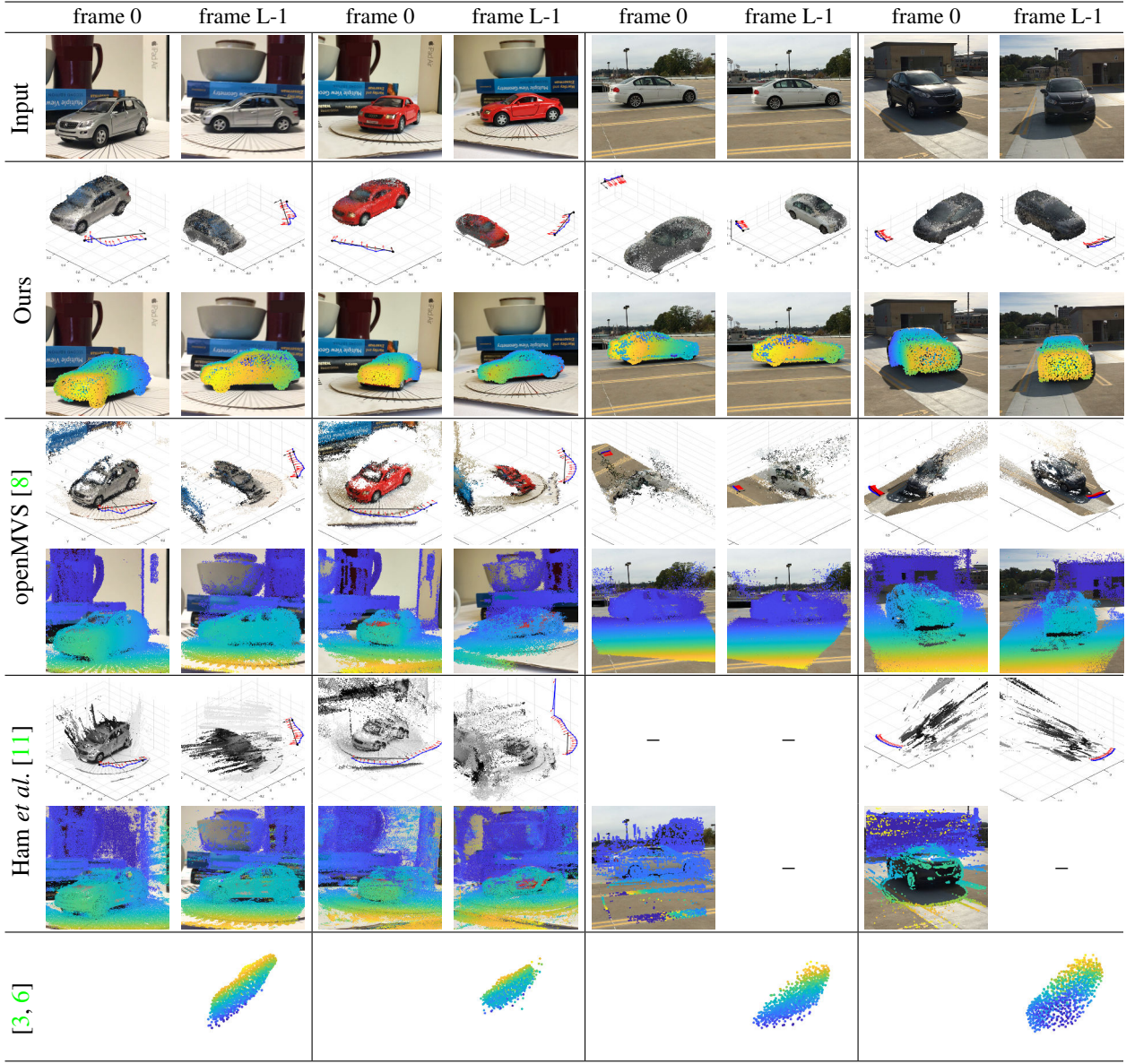

Figure 5: **Results of our methods on natural sequences, and comparison with openMVS [8], Ham *et al*. [11], and deep learning based methods [3, 6].** For openMVS [8] and Ham *et al*. [11] we align the cameras to the world coordinate frame (note that our method automatically produces camera poses in world frame), and we draw the aligned shape, estimated camera trajectory (blue) with orientation (red), annotated camera (black) of two frames (marked with black dot). We also project the shape with the estimated camera, and color the reprojected points with their inverse depths (brighter is closer). In the last row we show for each sample results from 3D-R2N2 [3] (left) and Fan *et al*. [6] (right).

results are reported in Table 1. We also show qualitative results on 4 natural sequences (2 model cars plus 2 real cars) in Fig. 5. Each sequence consists of $L = 91$ frames with roughly 30 degree of camera rotation and moderate translation. We show that we achieve comparable camera poses and denser inverse depths against openMVS [8] and Ham

*et al*. [11]. Additionally ours recovers semantic information including full 3D shape detached from the map, and again global cameras. For sequence 3 and 4 Ham *et al*. [11] gives degraded solution when it fails at camera tracking or deification due to little motion in frame 3 or significant lighting change in sequence 4. We attempt to give part of its results

|  | Rendered Sequences | | | | Natural Sequences | |
| --- | --- | --- | --- | --- | --- | --- |
|  | Depth Error | Density | Cam. Location Error | Cam. Orientation Error | Depth Error | Density |
| Zhu *et al.* | 0.0732 | 0.8543 | 0.0134 | 2.0254° | - | - |
| Ham *et al.* | 0.0682 | 0.4732 | **0.0045** | **0.9832°** | 0.0804 | 0.6558 |
| openMVS | 0.0731 | 0.5862 | 0.0098 | 1.3220° | 0.0798 | 0.6987 |
| Ours | **0.0627** | **0.9342** | 0.0102 | 1.2343° | **0.0756** | **0.9076** |

Table 1: **Quantitative comparison on both rendered and natural sequences.**

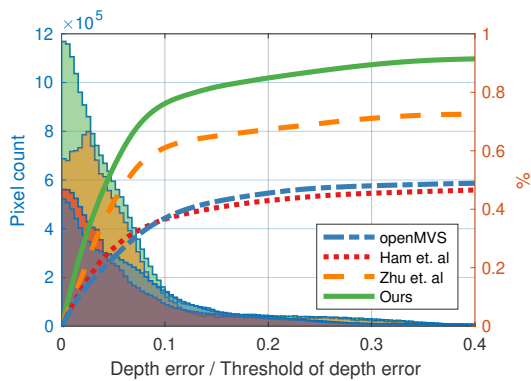

Figure 6: **Depth error on the rendered test set.** On the left axis shows the histogram of depth error distribution, and on the right axis gives the percentage of pixels under the threshold.

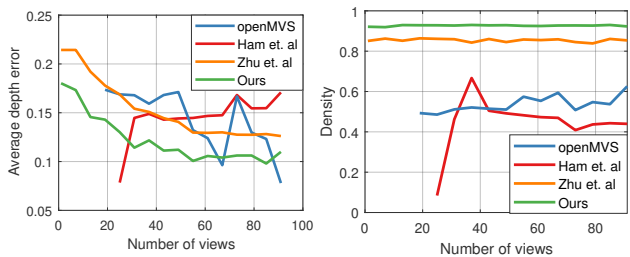

Figure 7: **Depth error on the rendered test set.** On the left axis shows the histogram of depth error distribution, and on the right axis gives the percentage of pixels under the threshold.

in the figure for readers' reference.

We also notice that learning based methods [6, 3] which are mostly trained on rendered images suffer from the domain gap in our test natural sequences.

## 4. Conclusion

In this paper we propose the method of semantic photometric bundle adjustment for object-centric shape reconstruction from natural sequence, which exerts geometric constraints over the camera pose as well as the full 3D shape generated by a learned semantic shape prior. We extensively evaluate our approach on both rendered and natural settings against both classic PBA methods and deep learning based methods, and prove that it is capable to produce dense full 3D shape in world coordinates, as well as depth maps of PBA-like quality.

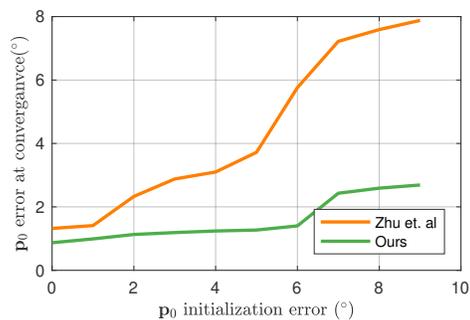

Figure 8: **Convergence analysis between ours and Zhu *et al.* [26].** Ours is more robust against initialization error in $p_0$.